\documentclass[runningheads]{llncs}

\usepackage{eccv}

\usepackage{eccvabbrv}

\usepackage{graphicx}
\usepackage{booktabs}
\usepackage{multirow}
\usepackage{marvosym}
\usepackage[accsupp]{axessibility}  

\usepackage{hyperref}

\usepackage{orcidlink}

\begin{document}

\title{Griffon: Spelling out All Object Locations at Any Granularity with Large Language Models} 

\titlerunning{Griffon}
\author{Yufei Zhan\inst{1,2}\orcidlink{0009-0002-1377-8519} \and Yousong Zhu\inst{1}\textsuperscript{,\Letter}\orcidlink{0000-0001-8544-410X}\and Zhiyang Chen\inst{1,2} \and Fan Yang\inst{1,3}\and Ming Tang\inst{1,2}\orcidlink{0000-0003-4976-3095} \and Jinqiao Wang\inst{1,2,3,4}\orcidlink{0000-0002-9118-2780}}

\authorrunning{Zhan et al.}

\institute{Foundation Model Research Center, Institute of Automation, \\Chinese Academy of Sciences, Beijing, China \and 
School of Artificial Intelligence, \\University of Chinese Academy of Sciences, Beijing, China \and 
Peng Cheng Laboratory, Shenzhen, China \and 
Wuhan AI Research, Wuhan, China\\
\email{\{zhanyufei2021,yangfan\_2022\}@ia.ac.cn}\\
\email{\{yousong.zhu,zhiyang.chen,tangm,jqwang\}@nlpr.ia.ac.cn}}

\maketitle
\def\thefootnote{}\footnotetext{\scriptsize Corresponding Author\Letter: Yousong Zhu.}
\begin{abstract}
  Replicating the innate human ability to detect all objects based on free-form texts at any granularity remains a formidable challenge for Large Vision Language Models (LVLMs). Current LVLMs are predominantly constrained to locate a single, pre-existing object. This limitation leads to a compromise in model design, necessitating the introduction of visual expert models or customized head structures. Beyond these constraints, our research uncovers LVLMs' capability for basic object perception, allowing them to accurately identify and locate objects of interest. Building on this insight, we introduce a novel Language-prompted Localization Dataset to fully unleash the capabilities of LVLMs in fine-grained object perception and precise location awareness. More importantly, we present Griffon, a purely LVLM-based baseline, which does not introduce any special tokens, expert models, or additional detection modules. It simply maintains a consistent structure with popular LVLMs by unifying data formats across various localization-related scenarios and is trained end-to-end through a well-designed pipeline. Comprehensive experiments demonstrate that Griffon not only achieves state-of-the-art performance on the fine-grained RefCOCO series and Flickr30K Entities but also approaches the capabilities of the expert model Faster RCNN on the detection benchmark MSCOCO. Data, codes, and models are released at \url{https://github.com/jefferyZhan/Griffon}.
\end{abstract}

\section{Introduction}
\label{sec:intro}
Large Language Models (LLMs)\cite{touvron2023llama,zheng2023judging} have shown state-of-the-art performances on a wide range of NLP tasks for capturing more nuanced relationships and contexts and better generalization from limited samples. Subsequently, Large Vision Language Models (LVLMs)\cite{instructblip, liu2023llava, zhu2023minigpt} have been naturally proposed as a general assistant to employ a LLM as the central processor to understand human intents and finish corresponding tasks. Though superior in various vision understanding tasks\cite{antol2015vqa, chen2015microsoft, johnson2017clevr}, they are limited in fine-grained object perception and spatial localization. Additionally, concerns have been raised about the occurrence of hallucinations\cite{liu2023mitigating} thought to be attributed to the lack of fine-grained supervision\cite{yin2023woodpecker}.

\begin{figure}[t]
  \centering
   \includegraphics[width=0.95\linewidth]{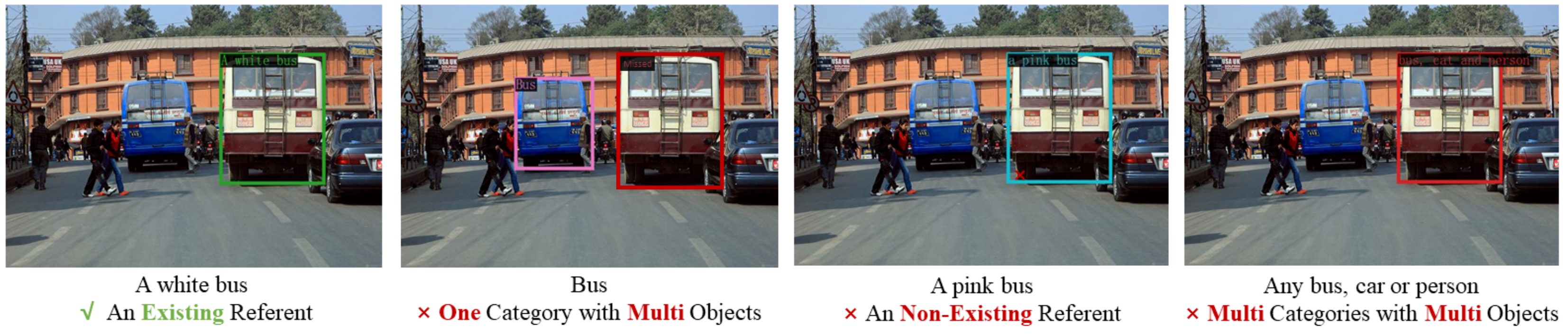}
   \caption{{\bf Four Types of Localization-Related Scenarios.} The overall localization task is partitioned into four scenarios based on the number of labels and the number of objects involved. Current LVLMs fail to refuse non-existing objects and detect multiple objects from one or multi-target descriptions (referents, categories, phrases \etc).}
   \label{fig:Compare}
\end{figure}

On the one hand, the limitation primarily stems from the LVLMs' weak understanding of multiple objects. As depicted in \cref{fig:Compare}, current popular LVLMs like Shikra\cite{chen2023shikra}, Qwen-VL\cite{Qwen-VL}, and KOSMOS-2\cite{kosmos-2}, equipped with basic object perception capabilities, are constrained to localize a single, pre-existing object and struggle in more complex scenarios. Despite increasing the volume to million level \cite{kosmos-2, Qwen-VL, xuan2023pink, you2023ferret}, the scenario diversity is still limited to the Referring Expression Comprehension (REC). In essence, the crux is not the quantity of data but the lack of comprehensive and well-arranged training benchmarks. 

On the other hand, several studies attempt either to invoke offline visual expert models\cite{liang2023taskmatrix, shen2023hugginggpt, zhao2023bubogpt} through LLMs or to integrate specific visual detection heads\cite{zhang2023gpt4roi,lai2023lisa, wang2023visionllm} to achieve fine-grained localization. The former is a scheduler-like approach, managing each visual expert model\cite{liu2023grounding, kirillov2023segment} through a specific Application Programming Interface (API). The latter generates specific task outputs through specialized head structures such as the detection head and segmentation head. While these models allow for nuanced predictions, they cannot enjoy a unified structure with LVLMs, increase the parameters exponentially, and fall short in utilizing the knowledge learned by LLMs to generalize to unseen objects. Therefore, the optimal solution lies in enhancing LVLMs with fine-grained perception and localization capabilities intrinsically, without relying on external aids.

In this paper, we introduce a pioneering language-prompted localization benchmark, advancing significantly in bridging the existing data gap in this domain. This benchmark consists of nearly 6M prompted pre-training data and 900K localization instruction-following data constructed with the help of GPT-4V\cite{openai2023gpt4}. It covers all four localization-related scenes and over 76K categories as diverse as possible in open-source datasets. It serves as a crucial foundation for LVLMs to comprehensively enhance their capabilities in simultaneously locating multiple objects in complex scenarios. Leveraging this benchmark, we present \textbf{Griffon}, an innovative, purely LVLM-based baseline localizing objects at any granularity according to the input texts. It is characterized by its streamlined architecture and unified input-output representation, which adopts a conversational design without any special token, priors or extra detection heads. The training of \textbf{Griffon} is strategically divided into two stages: Stage \MakeUppercase{\romannumeral1} focuses on the foundation pre-training and aims to enhance its fine-grained multiple object localization ability, and Stage \MakeUppercase{\romannumeral2} involves further instruction tuning for full scenarios, significantly improving the comprehension of users' intention. Additionally, we introduce a novel, training-free scoring mechanism to rank object outputs, enhancing the model's ability to prioritize more confident detections.

In order to validate the object perception capability with accurate localization of our framework, we conduct comprehensive experiments on public datasets. \textbf{Griffon} achieves the state-of-the-art results on RefCOCO series\cite{yu2016modeling, nagaraja2016modeling} and Flickr30K Entities \cite{plummer2015flickr30k} and the results on par with the expert model on detection dataset MSCOCO\cite{lin2014microsoft}. To summarize, the contributions are as follows:\begin{enumerate}
    \item We first introduce the Language-prompted Localization Dataset that consists of nearly 6M basic pre-training data and 900K instruction-following data, encompassing all four possible localization-related scenarios and over 76K object categories. 
    \item We present \textbf{Griffon}, a unified LVLM-based baseline without extra structures. It is capable of localizing all objects at any granularity based on free-form input texts.
    \item We have conducted comprehensive experiments on REC datasets RefCOCO series, phrase grounding dataset Flickr30K Entities, and object detection dataset MSCOCO 2017. \textbf{Griffon} achieves new state-of-the-art results on RefCOCO series and Flickr30K Entities and approaches the performance of the detection expert model.    
\end{enumerate}

\section{Related Work}
\subsection{Localization Models}
Traditional localization models usually refer to single-modality object detection or instance segmentation models. According to the backbone type, they can be divided into CNN-based detectors \cite{girshick2015fast,lin2017feature,ren2015faster,he2017mask,dai2016r, redmon2016you, duan2019centernet} and transformer-based detectors \cite{carion2020end,zhang2022dino,zhu2020deformable, chen2021pix2seq}, represented respectively by Faster R-CNN\cite{ren2015faster} and DETR\cite{carion2020end}. However, both of them are limited to close-set categories and cannot be generalized to novel categories. With the development of vision-language models \cite{radford2021learning}, open-vocabulary localization methods \cite{gu2021open, zareian2021open, zhou2022detecting} can locate objects of novel categories. Building on these methods, MDETR\cite{kamath2021mdetr}, GLIP\cite{zhang2022glipv2, li2021grounded} and Grounding DINO\cite{liu2023grounding} further convert localization tasks into a grounded vision-language task incorporating object detection, phrase grounding, and Referring Expression Comprehension tasks into a single open-set model. Unlike the above methods, our method makes predictions in an auto-regression manner without relying on the detection head and generalizes to different granularity localization tasks and object categories based on the LLM instead of the alignment between the region proposal and the text.

\subsection{Large Vision Language Models}
The evolution of LLMs \cite{touvron2023llama, chowdhery2022palm, zhang2022opt} to accommodate multimodal inputs is a burgeoning field of interest. To tune interleaved image-text data, Flagmingo\cite{alayrac2022flamingo} designed a visual perceiver that incorporates the visual information into the adapter layers of the LLM. On the contrary, BLIP-2\cite{li2023blip2} aligns visual embeddings with textual embeddings with a Q-Former. A fully connected layer is subsequently used to feed the projected visual embeddings to LLM. Following the inclusion of instruction tuning from NLP\cite{wang2022self, honovich2022unnatural}, models like Mini-GPT4\cite{zhu2023minigpt}, mPLUG-OWL\cite{ye2023mplugowl}, InstructBLIP\cite{instructblip} have been developed. They preserve the Q-Former structure and improve the interface performance with curated instruction-following data. LLaVA\cite{liu2023llava} differs by projecting visual encoder outputs to the LLM through a trainable fully connected layer, optimized with the LLaVA-150K instruction dataset, which is commonly used for instruction tuning. More evidence of LVLMs' remarkable visual language understanding powers and task-specific adaptability comes from the unveiling of GPT-4V\cite{openai2023gpt4}. Some LVLMs\cite{chen2023shikra, kosmos-2, Qwen-VL, xuan2023pink,you2023ferret} recently have introduced REC data into their training dataset, endowing them with basic object perception skills. Nonetheless, the precision in object localization and adaptability to complex scenarios remains a hurdle. To overcome these challenges, we present an exhaustive training benchmark and formulate a robust baseline purely based on LVLMs.

\subsection{Localization-equipped LVLMs}
To rival the detection performance of specialist models, the integration of visual expert models has been a prevalent tactic. HuggingGPT\cite{shen2023hugginggpt} enhances GPT's capabilities by connecting with a suite of visual expert models through APIs, allowing it to trigger a detection model upon recognizing an object detection prompt. AutoGPT\cite{yang2023auto} goes beyond by serving as an agent that can execute the post-processing tasks after detection. Further researches\cite{liang2023taskmatrix,nakano2021webgpt, zhao2023bubogpt} have expanded the range of APIs and refined their proficiency in specific domains. Another approach enriches the LVLM framework with dedicated detection heads\cite{zhang2023gpt4roi,lai2023lisa, wang2023visionllm, yan2023universal}. Accurate detection or segmentation results are produced by certain approaches\cite{lai2023lisa, yan2023universal} depending on the LLM predictions. In contrast, models such as VisionLLM\cite{wang2023visionllm} and GPT4RoI\cite{zhang2023gpt4roi} feed the output of the detection head to the LLM to enhance the detection performance or achieve particular goals. Unlike the above approaches, our methodology does not use any external specialized heads or expert models for detection. Instead, we enhance our model's fine-grained object perception and spatial localization capabilities through our meticulously crafted dataset and unified representation.

\begin{figure*}[t]
  \centering
  \includegraphics[width=0.95\linewidth]{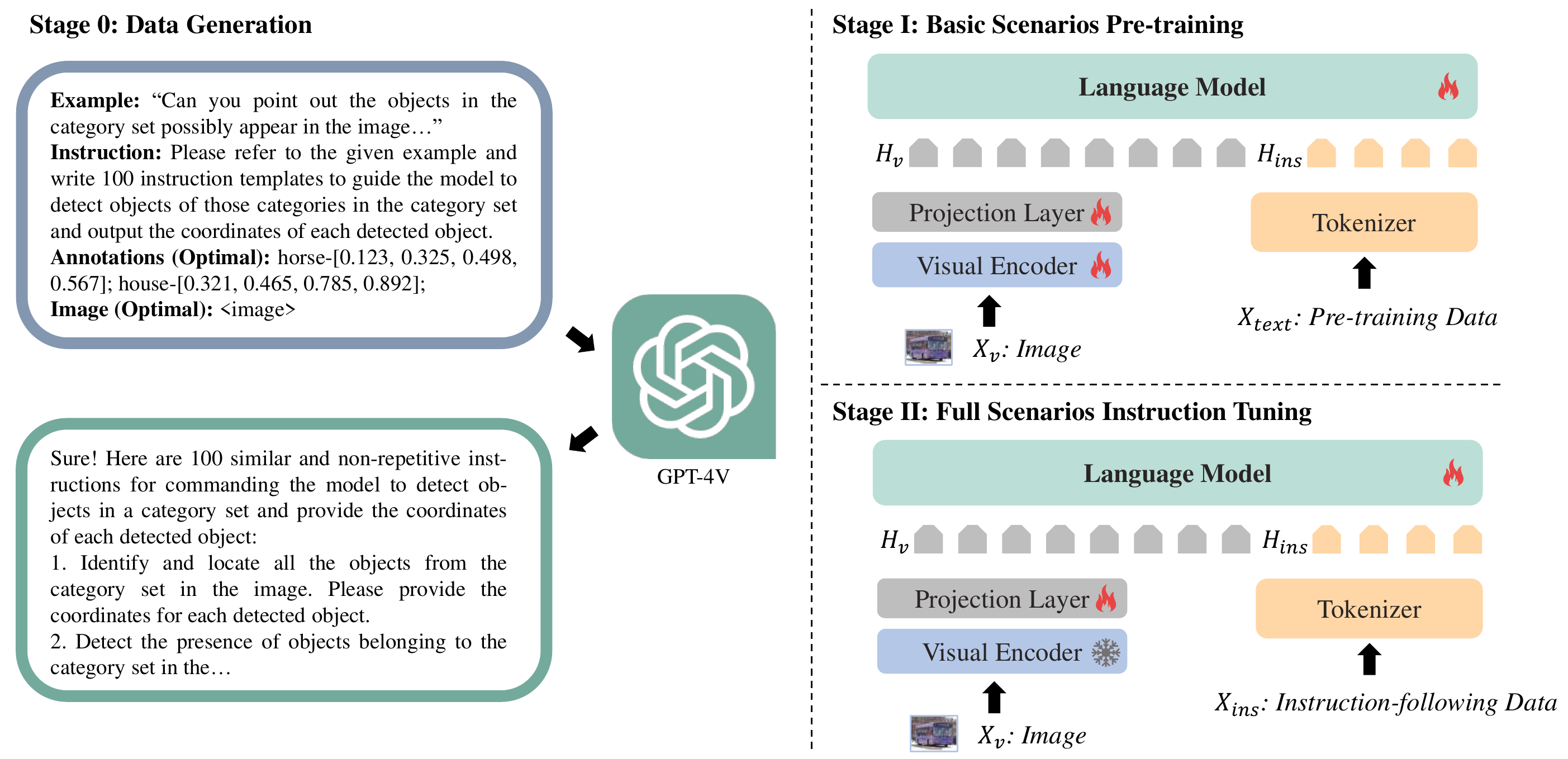}
   \caption{{\bf Data Generation and Training Procedure.} \textbf{Griffon} follows a progressive two-stage training pipeline with the built dataset in stage 0. In different stages, distinct modules of \textbf{Griffon} are trained. The red flame represents that this module is being trained at this stage, while the gray snowflake indicates the opposite.} 
   \label{fig: structure}
\end{figure*} 

\section{Methodology}
In this section, we start with our constructed novel Language-prompted Localization Dataset with scenarios analysis and detailed description of data construction in \cref{subsec: LPDD}. Leveraging this dataset, we present \textbf{Griffon}, a purely LVLM-based localization baseline. We detail the unified conversational input-output representation under a streamlined structure (\cref{subsec:unified}). And we further detail our two-stage training pipeline in \cref{subsec: training}. Finally, for the optimization of multi-object scenarios, we present the training-free scoring mechanism in \cref{subsec:score}.

\subsection{Language-prompted Localization Dataset}
\label{subsec: LPDD}
\subsubsection{Scenarios Analysis.}
\label{subsec: scene}
In the real world, a user may label the object with texts at different granularity. This granularity of texts can be clarified as fine-grained referents and coarse-grained categories. Current LVLMs\cite{chen2023shikra, kosmos-2} focus on the single referent localization and neglect the much more challenging multiple objects localization. Besides, there are scenarios where the target might not be present in the image. Hence, based on the quantity and type of labels specified, as depicted in \cref{fig:Compare}, language-prompted localization in the real-world setting can be categorized into four distinct scenarios:
\begin{itemize}
    \item Single Referent Localization (1 \vs 1): Models are required to distinguish the target from others and output the precise coordinates of that specific object.
    \item One Category with Multi Objects (1 \vs n): Upon receiving a category name or a descriptive phrase, the model locates and outputs the coordinates of all matching objects.
    \item An Non-existing Referent (None): If the input includes descriptions of nonexistent objects, the model indicates this by returning a 'None' output.
    \item Multi Categories with Multi Objects (n \vs n): When inputs comprise multiple descriptions, varying in both granularity and existence, the model identifies and outputs the coordinates of all discernible objects.
\end{itemize}

\subsubsection{Data Construction.}
\label{subsec: data construction}
Previous datasets\cite{liu2023llava, instructblip} employed by LVLMs are composed of two components, including pre-training data and instruction-following data. Datasets of different visual language tasks are merged into the pre-training data and then sampled and refined to build the instruction-following data. As for language-prompted localization, it's primarily related to the Referring Expression Comprehension (REC) above. Therefore, following the above dataset construction paradigm, we respectively build base pre-training data and instruction-following data for the Language-prompted Localization Dataset. We describe the main procedure as follows and details can be found in the Supplements.

\begin{figure*}[t]
  \centering
  \includegraphics[width=0.95\linewidth]{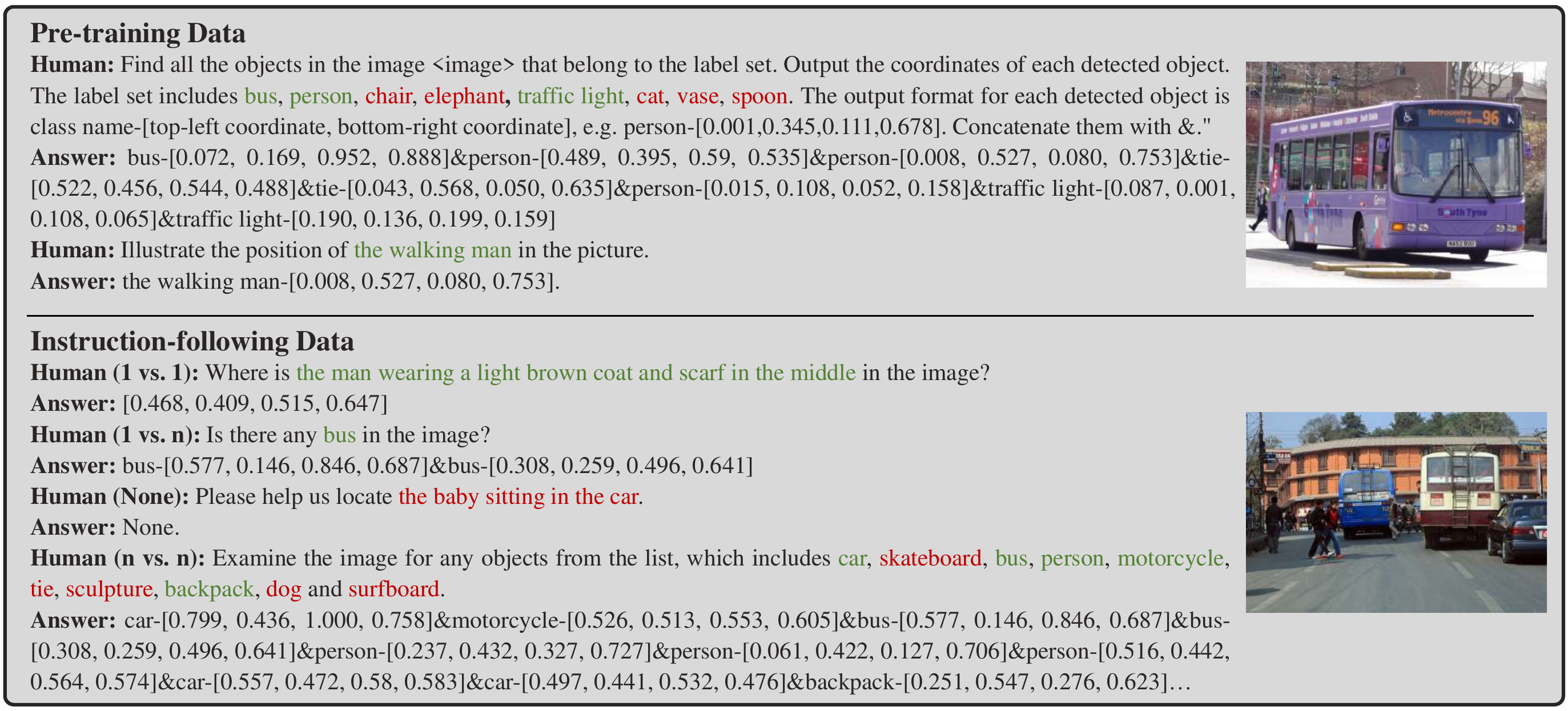}
   \caption{\textbf{Samples of Language-Prompted Localization Dataset.} All the images of the benchmark are collected and filtered from public datasets. Instructions are generated with the proposed method using GPT-4V\cite{openai2023gpt4}. The red indicates that this object does not exist in the image, while the green indicates the opposite.} 
   \label{fig: benchmark}
\end{figure*}

{\it Pre-training Data.} We first collect the public datasets of the REC task and object detection task such as Visual Genome\cite{krishna2017visual} and Objects365\cite{shao2019objects365}, which help with fine-grained discrimination and multiple object perception. After filtering images whose longest edge is less than 250 and whose image size is different from the size in the annotation, we totally collect 6M image-text samples, including nearly 4M in REC and 2M in object detection, respectively. As pure visual tasks, the annotations of these datasets only contain the labels without task prompts. Hence, we generate corresponding task templates to format the inputs with the help of GPT-4V\cite{openai2023gpt4}. As shown in the left part of \cref{fig: structure}, we first prompt GPT-4V with a template example, then ask GPT-4V to generate a large number of similar templates based on the requirements. As exampled in \cref{fig: benchmark}, each task template provides an interface, which can be flexibly replaced with other labels or phrases according to the ground truth during training. We craft 60 templates for each task, which can be randomly sampled.

{\it Instruction-following Data.} We construct the instruction data on the basis of the pre-training data according to the concluded four scenarios. Additionally, we incorporate the Flickr30K Entities \cite{plummer2015flickr30k} for the One Category with Multi Objects scenario and the LVIS dataset\cite{gupta2019lvis} for its ``negative categories'' annotations, which aims at building instruction samples of the Non-Existing Referent scenario. On one hand, we further generate more diverse instruction templates closer to the human daily expression with the help of GPT-4V and integrate the label with a randomly selected template to build instruction-image samples of different scenarios. On the other hand, for the case of Non-Existing Referent, we further mine negatives with a more fine-grained phrase using GPT-4V, adding extra attributes or descriptions, like ``a white cat'' or ``people wearing a scar''. It can deal with scenarios where the object exists but is described with wrong attributes. We collect nearly 900K unique image-text localization instruction samples in total and provide some examples of the final data in \cref{fig: benchmark}.

\subsection{Unified Representation With LVLMs}
\label{subsec:unified}
Locating all objects at any granularity through Large Language Models (LLMs) still remains to be explored in LVLMs. Thus, we develop \textbf{Griffon} based on the LLaVA\cite{liu2023llava} and propose a unified input-output representation. It generates predictions requiring only a single forward pass for multi-label targets without priors, structure modifications, and special tokens.

LLaVA is composed of three modules depicted in \cref{fig: structure}, including the visual encoder, projection layer, and LLM. We utilize the ViT-L/14 of CLIP\cite{radford2021learning} and LLama2-13B\cite{touvron2023llama} as our visual encoder and LLM, respectively, while the projection layer is a lightweight fully connection layer. The input image $X_v$ is first encoded to the visual feature $Z_v$ and then projected to word embedding space as $H_v$ through the projection layer. It shares the same dimension with the word embedding space and is concatenated with language embedding tokens $H_{ins}$, which are fed to the LLM to generate outputs.

Adopting the conversational style, the inputs of LLaVA are free-form texts ranging from commands to questions. Inheriting from the LLaVA, \textbf{Griffon} can also accept free-form inputs and further unifies inputs of different scenarios naturally. However, as LLaVA is aimed at visual language understanding tasks, it cannot handle the object coordinates exactly. Also, previous methods\cite{chen2023shikra, kosmos-2} can only output the textual coordinates of the single detected object and require repeated forward passes for multiple referents. Therefore, we present a unified output representation for multiple objects. For a detected object, it is represented with the top-left and bottom-right coordinates along with its label, formatted as ``label-[x1, y1, x2, y2]''. This can be adapted to multi-object scenarios seamlessly by concatenating all detected objects with ``\&'', without introducing connection placeholders\cite{Qwen-VL}. 

As for the representation of coordinates, there are mainly two types: extra special representation which creates new tokens to represent each coordinate in the axis, and numerical characters representation which directly uses numerals in natural languages. Shikra\cite{chen2023shikra} conducts a comparison experiment on these two representations on REC task and the conclusion is the numerical approach can achieve better performance. Also, as the embeddings of the extra tokens out of the original vocabulary need to be trained from scratch, it results in slower convergence. Hence, we employ numerical characters for normalized coordinates and set the precision to 0.001. This precision strikes a balance between reducing the localization error for small objects (area less than 32$\times$32 pixels) and the length of the sequence. 

\subsection{Progressive Two-Stage Training Pipeline}
\label{subsec: training}
Previous LVLMs equipped with localization capability add REC data to the vision and language alignment phrase and then are finetuned with designed localization-related instructions. By increasing the volume of data, this approach is feasible for tasks like REC that only output coordinates of the single object. However, when handling multiple objects of different granularities, requires the model to have a foundation understanding of the scene and the context of the entire image. Based on this insight and borrowing from the pre-training and instruction tuning paradigm from LLM, we propose the progressive two-stage training pipeline.

\subsubsection{Stage \MakeUppercase{\romannumeral1}: Basic Scenarios Pre-training.} The first stage is designed to enhance the multi-object perception capability and achieve localization in basic scenarios. We start by initialization with the weights from LLaVA to equip \textbf{Griffon} with the vision-language alignment capability and the instruction comprehension capability. Furthermore, we use the whole 6M pre-training data in the Language-prompted Localization Dataset for pre-training to create a foundation model that can accurately locate all objects in an image. To construct the input language $X_{ins}$, for each image $X_v$, a template will be sampled randomly from the all templates generated and integrated with the label for better generalization. Instead of only training the projector, we train the entire network. In this way, with fine-grained supervision, the details will be kept in the visual features, beneficial for small targets and fine-grained discrimination.

\subsubsection{Stage \MakeUppercase{\romannumeral2}: Full Scenarios Instruction Tuning.} To accustom to the full scenario, we conduct instruction tuning on the foundation model developed in the stage \MakeUppercase{\romannumeral2}. To be specific, following the first pre-training stage, the model has already developed the ability to localize objects based on free-form texts at any granularity. However, it fails to comprehend the instructions from specific scenarios and user intentions. Therefore, we further finetune the LLM and projection layer of \textbf{Griffon} with the 900K instruction data covering all four scenarios to refine the model's capabilities of comprehending the user's intention.

\subsection{Training-Free Confidence Scoring Mechanism}
\label{subsec:score}
In the object detection, expert models often predict an additional confidence score for each detected object to highlight high-quality bounding boxes. This approach mirrors the human cognitive process, which prioritizes attention towards objects of confidence based on existing knowledge. As a general optimization strategy, we present a training-free multi-object confidence scoring mechanism for localization with LVLMs. It generates an output sequence $X_a$ of length $L_a$ in an auto-regressive way, and the probability is computed by: 
\begin{equation}
p(X_{a}|X_v, X_{ins}) = \prod_{j=1}^{L_a} p_\theta(x_{a,j}| X_v, X_{ins}, X_{a,<j})
\end{equation}
where $\theta$ is the model parameters, $X_{ins}, X_{a,<j}$ are the instructions and answer tokens before the current prediction token $x_{a,j}$, respectively. At each step of the generation phrase until the $[EOS]$ token, we will get the conditional probability $p_\theta(x_{a,j}| X_v, X_{ins}, X_{a, <j})$ of $x_{a,j}$. For each predicted object label $R$, it consists of $L_r$ tokens. Supposing this label starts at $k^{th}$ token in the sequence, the conditional probability of generating this label can by computed according to Bayes' theorem by:
\begin{equation}
p(R|X_v, X_{ins}, x_{a, <k}) = \prod_{j=k}^{k+L_r} p_\theta(x_{a,j}|X_v, X_{ins}, X_{a, <j})
\end{equation}
where $X_v, X_{ins}$ are the visual and instruction inputs, and $x_{a, <k}$ are the answer tokens up to the $k^{th}$ token. This applies to the coordinates as well. With the conditional probability of each coordinate, we compute the localization confidence score for a bounding box starting from the $k^{th}$ token by:
\begin{equation}
p_{loc} = \prod_{coord \in \{x1, y1, x2, y2\}} p(coord|X_v, X_{ins}, X_{a, <k}),
\end{equation}
where $x1, y2, x2, y2$ are the top-left and bottom coordinates, respectively. 
The final confidence score for a detected object is the geometric mean of the label score and localization score, where $q=0.5$:
\begin{equation}
score = {p(R|X_v, X_{ins}, X_{a, <j})}^q \times {p_{loc}}^{1-q}.
\end{equation}

\section{Experiments}
\label{expr}

In this section, we first introduce the key implementation details briefly. Based on the implementation, We conduct quantitative experiments to evaluate the fine-grained object perception capability with spatial localization in three localization-related scenarios separately, including Single Referent, One Category with Multi Objects, and Multi Categories with Multi Objects scenarios. They correspond to REC, phrase grounding, and object detection tasks respectively. After that, ablation studies are provided to better comprehend the detailed options of \textbf{Griffon}. A qualitative analysis of all four localization-related scenarios is provided in the last.

\subsection{Implementation Details}
To strike a balance between the training efficiency and small objects performance, we increase the image input size to 448, applying the bilinear interpolation to the position embedding of the visual encoder. After the initialization, we adopt the burn-in strategy to warm up \textbf{Griffon} with a subset of REC data for 1 epoch. The learning rate is set to 2e-5 with a global batch size of 128. Then, \textbf{Griffon} is trained according to the designed pipeline. With the increase of data in the stage \MakeUppercase{\romannumeral1}, we align the batch size with the available computation resources, set to 256 with the learning rate linear scaled to 4e-5. In stage \MakeUppercase{\romannumeral2}, we decrease the batch size to 128 together with the learning rate to 2e-5. More details can be found in the Supplements.

\begin{table*}
    \centering
    \caption{Referring Expression Comprehension results with $ACC @ 0.5$. * indicates the result trained with only REC grounding data for a fair comparison with the above models.}
        \begin{tabular}{c|c|c|c|ccc|ccc|cc}
        \toprule
         \multirow{2}{*}{Type}& \multirow{2}{*}{Model} & Data& \multirow{2}{*}{Size} &\multicolumn{3}{c|}{RefCOCO} & \multicolumn{3}{c|}{RefCOCO+} & \multicolumn{2}{c}{RefCOCOg}\\
         && Vol. && val & test-A & test-B & val & test-A & test-B & val & test \\
         \midrule
         \multirow{4}{*}{\rotatebox{90}{Specialists}}& MDETR \cite{kamath2021mdetr} & \multirow{4}{*}{-} & \multirow{4}{*}{-} & 87.5 & 90.4 & 82.7 & 81.1 & 85.5 & 73.0 & 83.3 & 83.3\\
         & TransVG \cite{deng2021transvg} &  & & 81.0 & 82.7 & 78.4 & 64.8 & 70.7 & 56.9 & 68.7 & 67.7 \\
         & G-DINO-L \cite{liu2023grounding} &  & & 90.6 & 93.2 & 88.2 & 82.8 & 89.0  & 75.9 & 86.1 & 87.0\\
         & UNINEXT-L \cite{yan2023universal}&  & & 91.4 & 93.7 & 88.9 & 83.1 & 87.9  & 76.2 & 86.9 & 87.5\\
         \hline
         \multirow{7}{*}{\rotatebox{90}{Generalists}}&OFA-L \cite{wang2022ofa} & 10M & 512 &80.0 & 83.7 & 76.4 & 68.3 & 76.0 & 61.8 & 67.8 & 67.5 \\
         &Qwen-VL \cite{Qwen-VL}& 77M & 448& 89.4 & 92.3 & 85.3 & 83.1 & 88.3 & 77.2 & 85.6 & 85.5\\
         &PINK \cite{xuan2023pink}& 5M & 224 & 88.7 & 92.1 & 84.0 & 81.8 & 88.2 & 73.9 & 83.9 & 84.3 \\
         &Ferret-13B \cite{you2023ferret}& 8.7M & 336 & 89.5 & 92.4 & 84.4 & 82.8 & 88.1 & 75.2 & 85.8 & 86.3 \\
         &Shikra-13B \cite{chen2023shikra} & 4M & 224 &87.8 & 90.6 & 80.2 & 82.9 & 87.8 & 74.4 & 82.6 & 83.2\\
         &\textbf{Griffon-13B}& 7M & 448 & 89.4 &  92.5 &  84.6 & 83.3  &  88.4 &  76.0 &  85.1 &  86.1 \\
         &\textbf{Griffon-13B*}& 4M & 448 & \textbf{90.1} & \textbf{ 93.4} &  \textbf{86.1} & \textbf{84.8} & \textbf{90.5}  & \textbf{77.8}  &  \textbf{86.1} &  \textbf{87.2} \\
         \bottomrule
    \end{tabular}
    \label{tab:REC results}
\end{table*}

\subsection{Experiments on Referring Expression Comprehension}
Referring Expression Comprehension is aimed at the Single Referent scenario, reflecting the fine-gained determination capability with spatial localization. In \cref{tab:REC results}, we compare \textbf{Griffon} with a group of popular frameworks on RefCOCO\cite{yu2016modeling}, RefCOCO$+$\cite{yu2016modeling}, and RefCOCOg\cite{nagaraja2016modeling}. We mainly compare \textbf{Griffon} with the generalists based on LVLMs, because they share a similar structure and training approach. As these generalists are different in the scale of LLMs, input resolution, and the volume of data, we list the characteristics for comparison. Meanwhile, we also present the results of some experts to demonstrate the level \textbf{Griffon} has reached. As shown in \cref{tab:REC results}, \textbf{Griffon} surpasses the state-of-the-art model Qwen-VL under the same input resolution and with quite less training data, and also outperforms other models utilizing a similar amount of data on all metrics by a large margin. For fair comparison of the training pipeline we build, we also follow the previous models to only employ single referent data during pre-training. This makes the model focus on the dominant single referent localization, though the expression annotated might correspond to multiple objects. It is the crux of the previous high performance on REC tasks but failure in all multi-object scenarios for current generalists. In this setting, our model also achieves state-of-the-art results, proving the effectiveness of our pipeline.

\subsection{Experiments on Phrase Grounding}
Besides the Single Referent scenario, we present quantitative evaluations for the harder One Category with Multiple Objects scenario on the phrase grounding task. Existing LVLMs are typically limited to the Single Referent scenario, predicting only one box per phase, often missing out on multiple instances. In this way, they are only evaluated with the MERGED-BOXES protocol ignoring the single instance in the group the phrase refers to. \textbf{Griffon} excels in multi-object scenarios, seamlessly localizing both fine-grained and coarse-grained phases. We first evaluate this scenario on Flickr30K Entities \cite{plummer2015flickr30k} with both ANY-BOX protocol and MERGED-BOXES protocol. As depicted in the \cref{tab: results on pg}, \textbf{Griffon} achieves state-of-the-art on Flickr30K Entities, validating its success in multiple objects scenarios.

\begin{table}
    \centering
    \caption{Phrase grounding results on Flickr30K Entities\cite{plummer2015flickr30k}. ANY represents the ANY-BOX protocol, which focuses on the atomicity of each instance, while MERGED represents the MERGED-BOXES protocol evaluating whether the model finds all the referred objects with a merged box.}
    \begin{tabular}{c|c|cc|cc}
        \toprule
        \multirow{2}{*}{Type} & \multirow{2}{*}{Model} & \multicolumn{2}{c|}{ANY} & \multicolumn{2}{c}{MERGED} \\
        &&val & test & \;val & \;test\\
        \midrule
         \multirow{4}{*}{Specialists}&   BAN \cite{kim2018bilinear}    &    -  &    67.9  &   -   &-\\
         & DDPN \cite{yu2018rethinking} & - & - & 72.8 & 73.5 \\
         & VisualBert \cite{li2019visualbert} &   70.4   &   71.3   &   -   & -\\
         & MDETR \cite{kamath2021mdetr} &  82.5   &   83.4   &82.3 & 83.8\\
         \midrule
         \multirow{4}{*}{Generalists} & UniTAB \cite{yang2022unitab} & - & - & 78.8 & 79.6\\
         & Ferret-13B \cite{you2023ferret} & - & - & 81.1 & \textbf{84.8}\\
         & Shikra-13B \cite{chen2023shikra} &  - & - & 77.4 & 78.4 \\
         & \textbf{Griffon-13B} & \textbf{83.7} &\textbf{ 84.2} &   \textbf{82.0}   & 82.8\\
         \bottomrule
    \end{tabular}
    \label{tab: results on pg}
\end{table}

\subsection{Experiments on Object Detection}
Object detection exactly integrates all four scenarios into one task as there exist images containing one or no existing category for some images, and might be the hardest task. It can reflect the capabilities of dealing with complex scenarios. We conduct experiments on the widely-used object detection benchmark MSCOCO\cite{lin2014microsoft}. Since the existing popular LVLM methods cannot locate multiple objects of specified categories once, we evaluate Qwen-VL and Ferret on object detection tasks following the model design to ask one category at a time and iterate through all categories for each image. However, the $mAP$ and $AP_S$ of them are approximately 0, and the $AR_{100}$ of Qwen-VL is 3.7, which also validates the limitation of LVLMs we depict in the \cref{sec:intro} and \cref{fig:Compare}. We provide some visualizations of the test results in the Supplements for further illustration. Therefore, we don't include their results in the table and mainly compare our model with some basic expert models in \cref{tab:detection results}. As illustrated in \cref{tab:detection results}, \textbf{Griffon} is first capable of detecting multiple objects of various categories instead of 0 precision of existing LVLMs. It approaches the performance of the expert model Faster RCNN and outperforms Faster RCNN in $AP_S$ under the same resolution. With the above results, it can be concluded that \textbf{Griffon} is better than the previous generalist models with stronger localization and perception capability in both multi-object and single-object scenarios.

\begin{table*}
    \centering
    \caption{Object detection results on MSCOCO val2017 \cite{lin2014microsoft}. Griffon is the only LVLM capable of object detection, \ie detecting multiple objects of specified multiple categories once.}
    \begin{tabular}{c|c|cc|ccccccc}
        \toprule
         Type&Model& Size& Epochs & $mAP$&$AP_{50}$&$AP_{75}$&$AP_S$&$AP_M$&$AP_L$\\
         \midrule
         \multirow{3}{*}{Specialists}& Pix2Seq-R50 \cite{chen2021pix2seq}& 1333 & 300 & 43.0 & 61.0 & 45.6 & 25.1 & 46.9 & 59.4 \\
         & FRCNN-R50 \cite{ren2015faster}& 448 & 12 & 26.3 & 42.1 & 27.5 & 4.6 & 27.7 & 49.9 \\
         & DETR-DC5 \cite{zhang2022dino}& 1333 & 12 & 15.5 & 29.4 & 14.5 & 4.3 & 15.1 & 26.9 \\
         \midrule
         Generalists & \textbf{Griffon-13B} & 448 & 1 &24.8 &  40.6 &  25.1 & 5.9  &  25.5 &  48.7  \\
         \bottomrule
    \end{tabular}
    \label{tab:detection results}
\end{table*}

\subsection{Ablation Study}
In this section, we ablate on two key designs in \textbf{Griffon}. We conduct these experiments on MSCOCO val2017 \cite{lin2014microsoft} for demonstration.

\subsubsection{Unified Representation Order.} In \cref{subsec:unified}, we introduce the proposed unified representation which bridges the gap between single-referent scenarios and comprehensive scenarios with multiple objects for LVLMs. As \textbf{Griffon} generates outputs in the sequence-to-sequence manner, the relative positions of the coordinates and labels would affect the performance, especially for the multi-objects prediction. To compare the effectiveness of these two formats, we directly train the model with the pre-training data for 1 epoch for fast validation and evaluate them on the MSCOCO dataset. As shown in \cref{tab: order}, the label-first order surpasses the coord-first order on all metrics, with the mAP improved by +4.4\%. We believe the difference is caused by placing the label first can provide more uncertainty, as it corresponds to multiple different objects, to increase the number of predictions and further improve the recall rate. Therefore, we finally utilize the label-first format in \textbf{Griffon}.
\begin{table}
    \centering
    \caption{ Ablation study of unified representation order. The Coord-First order is ``[x1,y1,x2,y2]-label'', and the Label-First is the opposite.}
    \begin{tabular}{c|cccc}
        \toprule
        Order       &     mAP   & $AP_{50}$ & $AP_{75}$ & $AR_{100}$\\
        \midrule
        Coord-first &    17.8   &   28.9    &     17.9  &  30.5      \\
        Label-first &    22.2   &   35.4    &     22.6  &  33.0      \\
        \bottomrule
    \end{tabular}
    \label{tab: order}
\end{table}

\subsubsection{Confidence Scoring Mechanism.}
To optimize the multi-object predictions, we present the training-free confidence scoring mechanism. It produces the confidence of this prediction from both localization and recognition aspects with calculated conditional probability, detailed in \cref{subsec:score}. We validate the effectiveness of the proposed scoring mechanism in \cref{tab: Scoring mechanism}. The localization score can improve the multi-objects performance by +1.3\% mAP compared with the baseline. Composed of these two scores, the final score improves the mAP by 2.3\% overall. The result is quite evident because if removing the score, the evaluation degrades into a random process leading to mismatches. With the confidence score, the prediction with the higher score, \ie higher quality, will be matched with the ground truth. Therefore, it's better to utilize the training-free confidence scoring mechanism during inference.
\begin{table}
    \centering
    \caption{ Ablation study of Training-free Scoring Mechanism.}
    \begin{tabular}{cc|cccc}
        \toprule
         loc & label & mAP & $AP_{50}$ & $AP_{75}$ & $AR_{100}$\\
         \midrule
         $\times$ & $\times$ & 20.9 & 35.0& 20.7& 33.9\\
         $\checkmark$ & $\times$ & 22.2 & 35.4 & 22.6 & 33.0 \\
         $\checkmark$ & $\checkmark$ & 23.2 & 37.7 & 23.4 & 33.9\\
         \bottomrule
    \end{tabular}
    \label{tab: Scoring mechanism}
\end{table}

\subsection{Qualitative Analysis}
\begin{figure*}[t]
  \centering
   \includegraphics[width=0.9\linewidth]{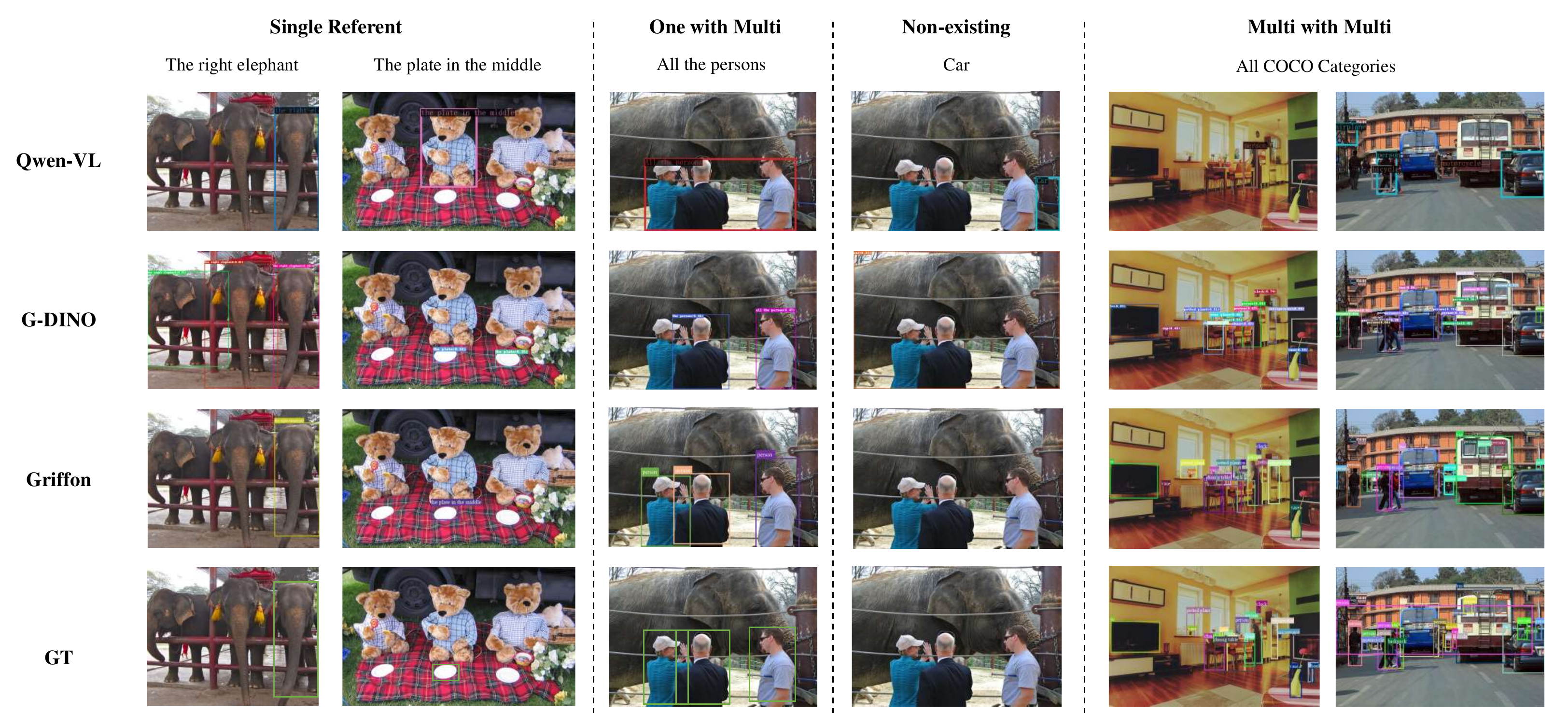}
   \caption{Visualization results of Qwen-VL\cite{Qwen-VL}, Grounding DINO\cite{liu2023grounding} and \textbf{Griffon} across all four scenarios. We use abbreviations to substitute the four scenarios.}
   \label{fig:visualization}
\end{figure*}

To gain more insights about \textbf{Griffon}, we present extensive visualization results in \cref{fig:visualization}. We qualitatively validate the effectiveness of \textbf{Griffon} by comparing it with the expert model Grounding DINO and the generalist model Qwen-VL in all possible localization scenarios. As shown in \cref{fig:visualization}, for Single Referent scenario, \textbf{Griffon} can not only localize objects of normal sizes but also small objects compared with Qwen-VL. Also, \textbf{Griffon} can refuse to output results when the input object is nonexistent and doesn't omit any objects in the One Category with Multi Objects scenario. It further approaches the performance of the expert model Grounding DINO in Multi Categories with Multi Objects scenario in the case of lower resolution (448 \vs 1333), accurately localizing the small objects and objects partially hidden. Therefore, our \textbf{Griffon} is capable of being applied in real-world scenarios requiring accurate localization.

\section{Conclusion}
\label{sec:conc}
In this paper, we present the first Language-prompted Localization Dataset across all possible localization-related scenarios. Leveraging this dataset, we establish a pure LVLM-based localization baseline, \textbf{Griffon}. It is capable of localizing objects based on the input texts at any granularity, without introducing any priors, detection modules, or special tokens. \textbf{Griffon} achieves the state-of-the-art results on the REC task and phrase grounding task and approaches the performance of visual expert model Faster RCNN on the detection benchmark. These results prove that a single LVLM can also do well in fine-grained visual perception tasks without external aids. We hope and believe our attempts at unified encoding for localization and the promising results achieved in object detection will facilitate the deep integration of visual tasks and vision-language tasks, offering potential utility to the broader community. As we mainly focus on localization and conduct experiments only in this domain, which might be a limitation, integrating more tasks into \textbf{Griffon} is our future direction.

\subsubsection{Acknowledgement.}{This work was supported by National Science and Technology Major Project (2021ZD0114600), National Natural Science Foundation of China (No.62276260, No.62176254) and in part by the Beijing Natural Science Foundation under Grant No.4244099. }

%
%
\bibliographystyle{splncs04}
\bibliography{main}

\end{document}